\documentclass{article}

\usepackage{microtype}
\usepackage{graphicx}
\usepackage{subcaption}
\usepackage{booktabs} 

\usepackage[table]{xcolor}

\usepackage{hyperref}

\usepackage[preprint]{icml2026}

\usepackage{amsmath}
\usepackage{amssymb}
\usepackage{mathtools}
\usepackage{amsthm}

\usepackage{bm}

\usepackage{makecell}

\usepackage[capitalize,noabbrev]{cleveref}

\theoremstyle{plain}

\theoremstyle{definition}

\theoremstyle{remark}

\usepackage[textsize=tiny]{todonotes}

\icmltitlerunning{Rethinking FID Through the Geometry of the Reference Dataset}

\begin{document}

\twocolumn[
  \icmltitle{Rethinking FID Through the Geometry of the Reference Dataset}



  \icmlsetsymbol{equal}{*}

  \begin{icmlauthorlist}
    \icmlauthor{Yunghee Lee}{comp}
    \icmlauthor{Byeonghyun Pak}{comp}
  \end{icmlauthorlist}

  \icmlaffiliation{comp}{Work done while at Agency for Defense Development, Daejeon, South Korea}

  \icmlcorrespondingauthor{Byeonghyun Pak\\}{bhpak@umd.edu}

  \icmlkeywords{Generative Models, Evaluation Metrics, FID}

  \vskip 0.3in
]



\printAffiliationsAndNotice{}  

\begin{abstract}
Fr\'echet Inception Distance (FID) is widely used to evaluate image generators, yet lower FID does not always correspond to better sample quality. 
We show that this mismatch depends in part on the geometry of the reference dataset. 
In a controlled study across six datasets, distributional density and effective rank significantly explain how FID changes as sample quality improves. 
Concentrated datasets tend to yield more favorable FID trends, whereas more dispersed datasets can make FID worsen despite better samples. 
Attribution to precision and recall and ablations with alternative feature spaces and distances support the same conclusion. 
These results suggest that distributional metrics should be interpreted together with the geometry of the reference dataset for more reliable benchmarking.
\end{abstract}

\section{Introduction}

Recent progress in image generation~\citep{saharia2022photorealistic, ramesh2022hierarchical, rombach2022high} has been accompanied by the widespread adoption of Fr\'echet Inception Distance (FID)~\citep{heusel2017gans} as a standard metric for evaluating generative models.
FID estimates the discrepancy between the distribution of Inception-v3~\citep{szegedy2016rethinking} features extracted from real images and that of generated images, and lower FID has therefore been widely interpreted as evidence of a better generator.
In practice, FID has become not only a reporting convention but also a benchmark target that shapes model development, hyperparameter tuning, and claims of progress across the literature~\citep{lu2025dpm, kim2025mixdit}.
This central role implicitly assumes that improvements in FID correspond to improvements in the qualities that matter for downstream or real-world use.

However, a growing body of evidence suggests that this assumption is unreliable.
\citet{choi2025enhanced} reported that allocating more computation per sample produces visibly sharper images, yet worsens FID on the COCO dataset~\citep{lin2014microsoft}.
\citet{lee2025tortoise} showed that tuning a generation hyperparameter to minimize FID can yield the worst per-sample quality as measured by ImageReward~\citep{xu2023imagereward}.
Likewise, \citet{jayasumana2024rethinking} demonstrated cases where stronger image distortions lead to better FID scores.
If FID simply measured generator quality, such adjustments should not consistently degrade the metric while improving perceptual quality, or vice versa.
These contradictions raise a broader concern: current benchmarking practice may fail to provide reliable or predictive insight into model behavior in realistic settings, even when the reported metric appears rigorous and standardized.

\begin{figure}[t]
  \centering
  \includegraphics[width=0.635\linewidth]{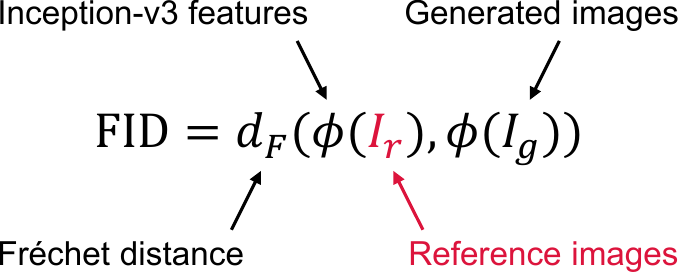}
  \vspace{5pt}
  \caption{
    Distributional metrics depend on four components.
    Prior work has studied the generator, feature extractor, and estimator; we investigate the reference dataset.}
    \vspace{-10pt}
  \label{fig:teaser}
\end{figure}

Prior work has mainly sought explanations in the fragility of Inception-v3~\citep{kynkaanniemi2022role, jayasumana2024rethinking, parmar2022aliased} and in the limitations of the Fr\'echet distance itself~\citep{chong2020effectively, lee2022trend}.
While these analyses are important, they leave one basic aspect of the metric underexplored: \textbf{the role of the reference dataset}.
Because FID is defined as a distance between the generated distribution and a chosen real-image reference distribution, its behavior necessarily depends on the structure of that reference set.
Yet the choice of reference dataset is rarely justified, and is often inherited from prior benchmarks~\citep{dehghani2021benchmark, otani2023toward}.

A dataset such as the CelebA-HQ dataset~\citep{karras2017progressive,xia2021tedigan} concentrates around a dominant visual mode, whereas the COCO dataset~\citep{lin2014microsoft} spans a far broader and more heterogeneous set of semantic categories; these datasets therefore occupy markedly different regions of feature space.
This observation motivates a fundamental question for theory-informed benchmarking: \emph{which characteristics of the reference dataset determine the properties that FID favors?}

In this paper, we investigate how the geometry of a reference dataset shapes the behavior of FID.
We characterize each dataset using two geometric properties: \textbf{distributional density}~\citep{loftsgaarden1965nonparametric} and \textbf{effective rank}~\citep{roy2007effective}.
We then test whether these properties moderate the effect of generated sample quality on FID.
Our results show that distributional density is a significant moderator of the quality-FID relationship.
For dispersed datasets, FID can \emph{worsen} even when per-sample quality is \emph{improved}, offering a potential explanation for contradictions noted in earlier studies~\citep{stein2023exposing}.
We further decompose FID into precision and recall~\citep{kynkaanniemi2019improved} and find that, for such datasets, FID is more strongly correlated with recall, which can improve even as sample quality worsens.
This dependence on dataset remains significant when replacing Inception-v3 with DINOv2~\citep{oquab2023dinov2} and when replacing Fr\'echet distance with MMD~\citep{gretton2012kernel} as in KID~\citep{binkowski2018demystifying}.
Therefore we argue that \textbf{distributional metrics should be interpreted together with reference dataset geometry}, and that understanding the reference dataset is essential for reliable benchmarking of generative models.

In summary, our contributions are as follows:
\begin{itemize}
  \item We present a controlled empirical study showing that reference dataset geometry, quantified by distributional density and effective rank, significantly moderates FID.
  \item We perform ablations across alternative feature spaces (DINOv2) and alternative kernel-based metrics (KID), showing that the observed effect is not specific to Inception-v3 features or the Fr\'echet estimator.
  \item We provide practical guidance for evaluating distributional metrics: use FID with concentrated datasets, or report it together with the geometric descriptors.
\end{itemize}

\section{Method}

\subsection{Geometric Properties of Dataset}
\label{subsec:geometric_properties}

We summarize each reference dataset with two scalar descriptors of its feature distribution.
Both descriptors are computed in the same feature space as the metric.

\paragraph{Distributional Density.}
Distributional density measures how tightly the reference distribution concentrates around its own points.
A high value means the samples are clustered and close to each other; a low value means they are scattered and far away.

A standard estimator of concentration is the kNN density \citep{loftsgaarden1965nonparametric}, $\hat p(x) \propto d_k(x)^{-D}$, where $d_k(x)$ denotes the Euclidean distance from $x$ to its $k$-th nearest neighbor and $D$ is the feature dimension.
However, in the Inception-v3 feature space with $D=2048$, the exponent makes $\hat p$ impractical as a dataset-level summary, as its values span tens of orders of magnitude across datasets.
To obtain a more stable descriptor, we take the logarithm and then average over the reference set, reducing the pointwise estimate to a single scalar.
The resulting descriptor, which we call the \emph{mean kNN log-density}, is defined as
\begin{equation}
  \langle -\log d_k \rangle = \frac{1}{n} \sum_{i=1}^{n} -\log d_k(x_i),
\end{equation}
where $n$ is the number of reference points.
We use $k=80$ throughout this paper.

\paragraph{Effective Rank.}
Effective rank measures how many linear directions the feature distribution spreads over.
A high value means the support occupies many principal components of the feature space; a low value means it collapses onto a few.
The effective rank of a matrix $A$ is~\citep{roy2007effective}
\begin{equation}
  \mathrm{erank} (A) = \exp \left( H(\bm{\sigma} / \|\bm{\sigma}\|_1) \right),
\end{equation}
where $\bm{\sigma}$ is the vector of singular values of $A$, $\|\bm{\sigma}\|_1$ is the L1 norm of $\bm{\sigma}$, and $H(\bm{p}) = -\sum_k p_k \log p_k$ is the Shannon entropy~\citep{shannon1948mathematical}.
$\mathrm{erank}(A)$ equals $\mathrm{rank}(A)$ when nonzero singular values are all equal, which gives it the interpretation of weighted dimensionality.
We use $\mathrm{erank}(A)$ as the descriptor where $A$ is the centered feature matrix.

\begin{table}[t]
\caption{
  Distributional density $\langle -\log d_k \rangle$ and effective rank $\mathrm{erank}(A)$ of the six reference datasets in the Inception-v3 feature space. 
}
\label{tab:geometry}
\begin{center}
\begin{small}
\begin{tabular}{lcc}
\toprule
Dataset & $\langle -\log d_k \rangle$ & $\mathrm{erank}(A)$ \\
\midrule
FFHQ \citep{karras2019style}      & $-2.48$ & $1243$ \\
CelebA-HQ \citep{karras2017progressive} & $-2.36$ & $1220$ \\
MJHQ-30K \citep{li2024playground}  & $-2.74$ & $1341$ \\
ImageNet \citep{deng2009imagenet}  & $-2.68$ & $1431$ \\
Flickr30K \citep{young2014image} & $-2.80$ & $1341$ \\
COCO \citep{lin2014microsoft}      & $-2.67$ & $1337$ \\
\bottomrule
\end{tabular}
\end{small}
\end{center}
\end{table}

Table~\ref{tab:geometry} reports both descriptors on the six reference datasets used in this study.
The face datasets, FFHQ and CelebA-HQ, score highest on distributional density and lowest on effective rank, consistent with their single-domain structure.
The open-domain datasets, ImageNet, Flickr30K, and COCO, score in the opposite direction.

\begin{figure*}[t]
  \centering
  \includegraphics[width=\linewidth]{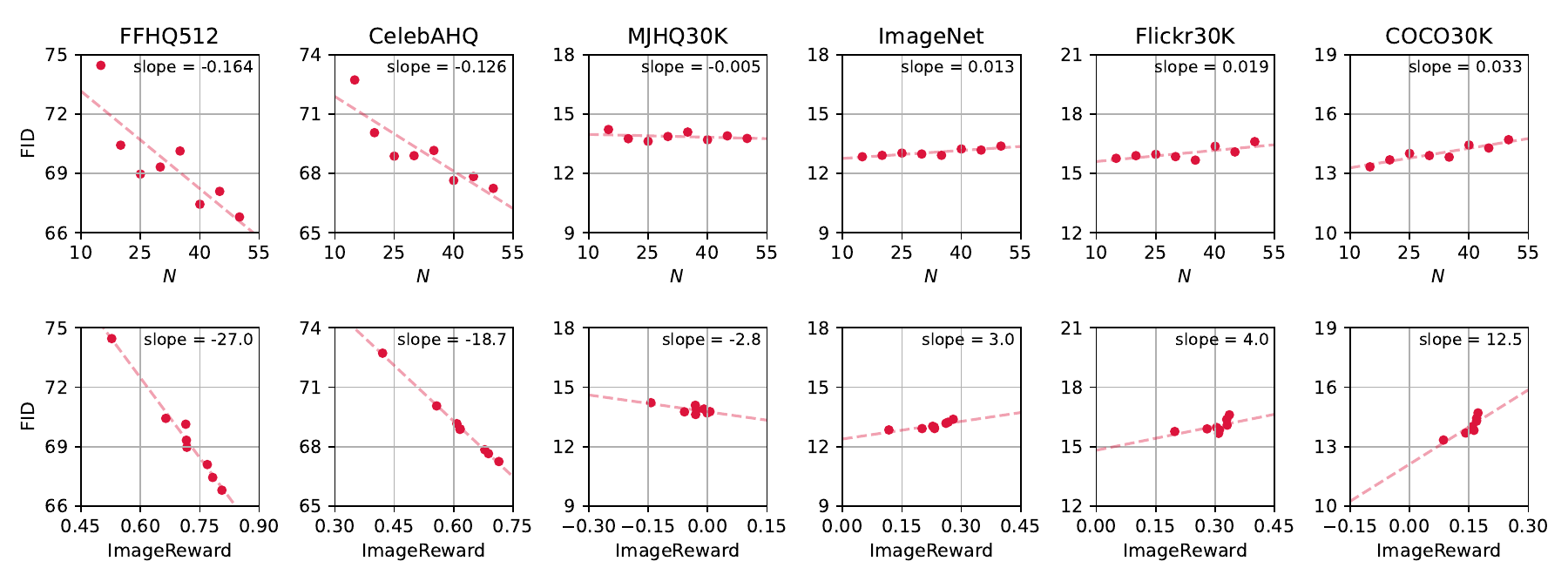}
  \caption{
    FID across six reference datasets under a fixed generator and a sweep over the number of denoising steps $N$.
    Top row: FID as a function of $N$.
    Bottom row: FID as a function of ImageReward over the same sweep.
    Although image quality generally improves with more denoising steps, FID exhibits dataset-dependent trends and may either increase or decrease depending on the reference dataset.
    }
  \label{fig:tradeoff}
  \vspace{-2mm}
\end{figure*}

\subsection{Statistical Analysis}

We investigate how FID behavior depends on the reference dataset through statistical analysis.
In particular, we focus on how FID fragility varies across reference datasets.
We ask: \emph{for a given dataset, how does FID change as per-sample quality improves?}
To answer this, we introduce an independent variable $X$ for sample quality and a dependent variable $Y$ for FID, and use the slope of a linear regression model to quantify their relationship.

\paragraph{Omnibus Test.}
We first test whether the slopes vary significantly across datasets.
The null hypothesis $H_0$ is that \emph{the slopes do not vary by dataset}, whereas the alternative hypothesis $H_1$ is that \emph{the slopes vary significantly across datasets}.
To test $H_0$, we fit a two-level hierarchical linear model~\citep{raudenbush2002hierarchical}.
Using a likelihood ratio test~\citep{stram1994variance}, we compute a test statistic $D$, which follows a mixed $\chi^2$ distribution under $H_0$.
We report $D$ and its corresponding $p$-value for testing $H_0$.
Additional details are provided in Appendix~\ref{sec:stats}.

\paragraph{Moderation Test.}
We next test whether the geometric descriptors moderate the slope.
To do so, we introduce a covariate $Z$, representing a geometric descriptor, into the hierarchical linear model and test for a cross-level interaction such that the slope varies with $Z$.
The null hypothesis $H_0$ is that \emph{the slope does not depend on $Z$}, whereas the alternative hypothesis $H_1$ is that \emph{the slope depends on $Z$}.
We report the cross-level interaction coefficient $\gamma_{11}$, its corresponding $p$-value from the Wald test~\citep{wald1943tests}, and $R^2_{\text{slope}}$, the proportion of slope variance explained by $Z$.
Additional details are provided in Appendix~\ref{sec:stats}.

\subsection{Experimental Design}
\label{subsec:design}

We analyze six reference datasets spanning a spectrum from concentrated to dispersed feature distributions: FFHQ and CelebA-HQ on the concentrated side, MJHQ-30K, ImageNet, Flickr30K, and COCO on the dispersed side.
For each dataset, we generate samples with Stable Diffusion 1.5~\citep{rombach2022high} using DDIM~\citep{song2020denoising} at $512\times512$, classifier-free guidance scale $7.5$, and a fixed random seed per prompt, while sweeping the number of denoising steps $N \in \{15, 20, 25, 30, 35, 40, 45, 50\}$.
Prompt sources per dataset are listed in Appendix~\ref{sec:datasets}.
For each pair of dataset and $N$ we generate one image per reference example, so the generated set matches the reference set in both size and conditioning.
We evaluate each set using FID, with KID and FD$_{\text{DINOv2}}$~\citep{stein2023exposing} as ablations, ImageReward~\citep{xu2023imagereward} as a proxy for perceptual quality, and precision and recall~\citep{kynkaanniemi2019improved} as diagnostic measures.

\section{Results}

\cref{fig:tradeoff} compares FID with per-sample quality across reference datasets.
The relationship between FID and sample quality exhibits distinct trends across datasets.
On CelebA-HQ and FFHQ, the slope is negative: FID decreases as the number of denoising steps increases and ImageReward improves.
By contrast, on COCO, Flickr30K, and ImageNet, the slope is positive.
MJHQ-30K lies between these two regimes.
These results show that although per-sample quality generally improves with more inference steps, FID may either increase or decrease depending on the reference dataset.
This suggests that FID does not reflect a fixed notion of generator quality, but rather favors different properties depending on the geometry of the reference dataset.

\subsection{Statistical Analysis}

\paragraph{Omnibus Test.}
We first test whether the slope of the relationship between sample quality and FID varies across datasets.
Using $X=N$ and $Y=\text{FID}$, the omnibus test rejects the null hypothesis that all datasets share a common slope ($D = 44.3$, $p<.001$).
We obtain the same conclusion when using $X=\text{ImageReward}$ and $Y=\text{FID}$ ($D = 90.9$, $p<.001$).
These results indicate that the slope differences in \cref{fig:tradeoff} reflect systematic differences in how FID responds to sample quality across datasets.

\begin{table}[t]
\caption{
  Statistical test results. IR stands for ImageReward. For each choice of $X$, the \colorbox{blue!10}{omnibus test} result ($D$ and $ p$) is reported first, followed by the moderation test results ($\gamma_{11}$,$ p$, and $ R^2_\text{slope}$).
}
\label{tab:moderation}
\begin{center}
\begin{small}
\begin{tabular}{cccccc}
\toprule
$X$ & $Y$ & $Z$ & $D$ or $\gamma_{11}$ & $p$ & $R^2_\text{slope}$ \\
\midrule
\rowcolor{blue!10}
$N$ & FID & - & $44.3$ & $<.001$ & - \\
$N$ & FID & $\langle -\log d_{k} \rangle$ & $-0.0323$ & $<.001$ & $0.707$ \\
$N$ & FID & $\mathrm{erank}(A)$ & $0.0314$ & $.002$ & $0.661$ \\
\midrule
\rowcolor{blue!10}
IR & FID & - & $90.9$ & $<.001$ & - \\
IR & FID & $\langle -\log d_{k} \rangle$ & $-0.120$ & $.007$ & $0.548$  \\ 
IR & FID & $\mathrm{erank}(A)$ & $0.119$ & $.010$ & $0.530$\\
\bottomrule
\end{tabular}
\end{small}
\end{center}
\end{table}

\begin{table}[t]
\caption{
    Coefficient of determination ($R^2$) of FID with precision and recall for each dataset.
    The larger value is highlighted in \textbf{bold}.
}
\label{tab:r_squared}
\begin{center}
\begin{small}
\begin{tabular}{lcc}
\toprule
Dataset & $R^2(\text{Precision, FID})$ & $R^2(\text{Recall, FID})$ \\
\midrule
FFHQ       & \textbf{0.989} & 0.672 \\
CelebA-HQ  & \textbf{0.951} & 0.001 \\
MJHQ-30K   & \textbf{0.734} & 0.025 \\
ImageNet   & 0.690 & \textbf{0.949} \\
Flickr30K  & 0.314 & \textbf{0.850} \\
COCO       & 0.676 & \textbf{0.833} \\
\bottomrule
\end{tabular}
\end{small}
\end{center}
\vspace{-3mm}
\end{table}

\paragraph{Moderation Test on Distributional Density.}
We next test whether distributional density explains why the slope differs across datasets; the results are reported in \Cref{tab:moderation}.
Using $X=N$, $Y=\text{FID}$, and $Z=\langle -\log d_{k} \rangle$, we find a significant negative cross-level interaction with $R^2_{\text{slope}}=0.707$.
This indicates that distributional density explains 70.7\% of the between-dataset variation in slopes.
Moreover, the negative sign of $\gamma_{11}$ indicates that denser reference datasets tend to have smaller, or more negative, $N$-to-FID slopes.
These results suggest that FID is more likely to improve with additional inference steps when the reference distribution is dense.
The same pattern holds with $X=\text{ImageReward}$, indicating that distributional density moderates the relationship between sample quality and FID.

\paragraph{Moderation Test on Effective Rank.}
We then test whether effective rank similarly explains the slope differences across datasets, with results also reported in \cref{tab:moderation}.
Using the same model with $Z$ defined as effective rank, we find a significant positive cross-level interaction ($\gamma_{11}=0.0314$, $p=.002$) with $R^2_{\text{slope}}=0.661$.
This indicates that effective rank explains 66.1\% of the between-dataset variation in slopes.
The positive sign of $\gamma_{11}$ indicates that datasets with higher effective rank tend to have larger $N$-to-FID slopes.
The same pattern holds with $X=\text{ImageReward}$.
In turn, this suggests that reference datasets spanning many directions in feature space are more likely to make FID worsen as sample quality improves.

\subsection{Attribution to Precision and Recall}
We further decompose FID into precision and recall to examine which component more strongly explains the observed variation in FID.
Precision measures the fidelity of generated images to the real data distribution, whereas recall measures their diversity, that is, how well the generated samples cover that distribution.
To assess which component is more strongly associated with FID, we compute the $R^2$ of separate ordinary least squares fits for each dataset.
The results are summarized in \cref{tab:r_squared}.
For concentrated datasets such as FFHQ and CelebA-HQ, FID is more strongly correlated with precision.
By contrast, for dispersed datasets such as ImageNet, Flickr30K, and COCO, FID is more strongly correlated with recall.
This suggests that when the reference dataset is dispersed, FID tends to reward coverage of the real distribution more than sample-level fidelity.
In other words, FID can \emph{worsen} even when precision \emph{improves}, if recall \emph{declines} as per-sample quality increases.
Viewed from this diagnostic perspective, the result supports our main finding that the behavior of FID depends systematically on reference dataset geometry.

\subsection{Ablation Study}

\begin{table}[t]
\caption{
    Ablation study results.
    For each alternative metric, the first row reports the \colorbox{blue!10}{omnibus test} results ($D$ and $p$), followed by moderation test results ($\gamma_{11}$, $p$, and $R^2_\text{slope}$).
}
\label{tab:ablation}
\begin{center}
\begin{small}

\begin{tabular}{cccccc}
\toprule
$X$ & $Y$ & $Z$ & $D$ or $\gamma_{11}$ & $p$ & $R^2_\text{slope}$ \\
\midrule
\rowcolor{blue!10}
$N$ & KID & - & $46.4$ & $<.001$ & - \\
$N$ & KID & $\langle -\log d_{k} \rangle$ & $-0.0343$ & $<.001$ & $0.763$ \\
$N$ & KID & $\mathrm{erank}(A)$ & $0.0315$ & $.005$ & $0.596$ \\
\midrule
\rowcolor{blue!10}
$N$ & FD$_\text{DINOv2}$ & - & $26.6$ & $<.001$ & - \\
$N$ & FD$_\text{DINOv2}$ & $\langle -\log d_{k} \rangle$ & $-0.0108$ & $<.001$ & $0.827$  \\ 
$N$ & FD$_\text{DINOv2}$ & $\mathrm{erank}(A)$ & $0.0110$ & $<.001$ & $0.837$\\
\bottomrule
\end{tabular}
\end{small}
\end{center}
\vspace{-3mm}
\end{table}

We perform an ablation study on other components of FID.
Specifically, we replace the Fr\'echet distance with kernel MMD to obtain KID, and replace Inception-v3 with DINOv2 to obtain FD$_\text{DINOv2}$.
\cref{tab:ablation} summarizes the results of the omnibus and moderation tests using these alternative distributional metrics.
In all cases, the omnibus test rejects $H_0$, indicating that the slope remains dataset-dependent regardless of the choice of distance or feature space.
This suggests that prior explanations focusing only on the distance measure or feature extractor do not fully explain the fragility of FID.
Moreover, in the moderation tests, the geometric descriptors explain even more slope variance with DINOv2 features than with Inception-v3 features, indicating that the observed effect is not driven by known limitations of Inception-v3.
These results show that our geometric descriptors generalize beyond FID and remain informative across alternative distributional metrics.
\section{Conclusion}

In this paper, we showed that the behavior of FID systematically depends on the geometry of the reference dataset.
Across 6 datasets, distributional density and effective rank significantly explained how FID changes as sample quality improves.
Concentrated datasets made FID more aligned with precision, whereas dispersed datasets made it more aligned with recall.
These findings remained robust across alternative feature spaces and distance measures.
Therefore, distributional distance metrics should be interpreted together with the geometry of the reference dataset.
When using FID as an evaluation criterion for generative models, we suggest using concentrated reference datasets such as FFHQ.
Conversely, for open-domain settings, we recommend reporting FID together with the geometric descriptors of the reference dataset.


\section*{Acknowledgements}

We sincerely thank Byeongju Woo and Hoseong Kim for their constructive discussions and support.
We also appreciate Sol Park and Minkyu Song for providing insightful feedback.
This work was supported by the Agency For Defense Development Grant Funded by the Korean Government (912A45701).

\section*{Impact Statement}


This paper presents work whose goal is to advance the field of Machine
Learning. There are many potential societal consequences of our work, none
which we feel must be specifically highlighted here.



\bibliography{main}
\bibliographystyle{icml2026}

\newpage
\appendix
\onecolumn

\section{Additional Details on Datasets}
\label{sec:datasets}

\begin{table}[t]
\caption{
  Details of datasets used for experiments.
}
\label{tab:datasets}
\begin{center}
\begin{small}
\begin{tabular}{lcccc}
\toprule
Dataset & Image count & Caption source & HuggingFace identifier \\
\midrule
FFHQ       & 70,000 & Auto-generated     & \texttt{Ryan-sjtu/ffhq512-caption} \\
CelebA-HQ  & 29,987 & Auto-generated     & \texttt{oftverse/control-celeba-hq} \\
MJHQ-30K   & 30,000 & Midjourney prompts & \texttt{xingjianleng/mjhq30k} \\
ImageNet   & 50,000 & Class names        & \texttt{BenSchneider/imagenet-val} \\
Flickr30K  & 31,014 & Human-annotated    & \texttt{nlphuji/flickr30k} \\
COCO       & 30,000 & Human-annotated    & \texttt{sayakpaul/coco-30-val-2014} \\
\bottomrule
\end{tabular}
\end{small}
\end{center}
\end{table}

\cref{tab:datasets} shows additional details of the datasets used for experiments.
All datasets were downloaded using the HuggingFace \texttt{datasets} library~\citep{lhoest2021datasets}.

\section{Additional Details on Statistical Tests}
\label{sec:stats}

In this section, we cover the details of statistical tests performed. 

\paragraph{Omnibus Test.}
We use a 2-level hierarchical linear model~\citep{raudenbush2002hierarchical} to find out if there is a significant difference between slopes across datasets.
Let $X_{ij}$ be the $X$ value where $i$ is the index for individual observation for each $N$ and $j$ is the index for datasets.
Similarly, $Y_{ij}$ is the $Y$ value for observation $i$ in dataset $j$.
Corresponding to the null hypothesis $H_0$ such that there is no difference of slopes, we fit a random intercepts only model 
\begin{align}
    Y_{ij} &= \beta_{0j} + \beta_{1j} X_{ij} + \epsilon_{ij},
    & \epsilon_{ij} &\sim \mathcal{N}(0, \sigma^2), \\
    \beta_{0j} &= \gamma_{00} + u_{0j},  
    & u_{0j} &\sim \mathcal{N}(0, \tau_{00}), \\
    \beta_{1j} &= \gamma_{10}
\end{align}
for the parameters $\gamma_{00}$, $\gamma_{10}$, and $\tau_{00}$.
For the alternative hypothesis $H_1$, we fit a random intercepts and slopes model
\begin{align}
    Y_{ij} &= \beta_{0j} + \beta_{1j} X_{ij} + \epsilon_{ij},
    & \epsilon_{ij} &\sim \mathcal{N}(0, \sigma^2), \\
    \beta_{0j} &= \gamma_{00} + u_{0j}, \\
    \beta_{1j} &= \gamma_{10} + u_{1j}, 
    & \begin{bmatrix} u_{0j} \\ u_{1j} \end{bmatrix} &\sim \mathcal{N}\left(
        \begin{bmatrix} 0 \\ 0 \end{bmatrix},
        \begin{bmatrix}
            \tau_{00} & \tau_{01} \\ \tau_{01} & \tau_{11}
        \end{bmatrix}
    \right)
\end{align}
for the parameters $\gamma_{00}$, $\gamma_{10}$, $\tau_{00}$, $\tau_{01}$, and $\tau_{11}$.
Then we perform a likelihood ratio test (LRT) between these two models. 
Let the likelihood of each model be $L_0$ and $L_1$.
The test statistic $D = -2 (\log L_0 - \log L_1)$ is known to follow a 50:50 mixture of $\chi_1^2$ and $\chi_2^2$ distributions under $H_0$~\citep{stram1994variance}.
We report the test statistic $D$ and the corresponding $p$-value.

\paragraph{Moderation Test.}

We use a similar model to find out if a geometric descriptor covariate $Z$ moderates the relationship between $X$ and $Y$.
Let $Z_j$ be the value of $Z$ for dataset $j$.
We fit the model
\begin{align}
    Y_{ij} &= \beta_{0j} + \beta_{1j} X_{ij} + \epsilon_{ij},
    & \epsilon_{ij} &\sim \mathcal{N}(0, \sigma^2), \\
    \beta_{0j} &= \gamma_{00} + \gamma_{01} Z_j + u_{0j}, \\
    \beta_{1j} &= \gamma_{10} + \gamma_{11} Z_j + u_{1j}, 
    & \begin{bmatrix} u_{0j} \\ u_{1j} \end{bmatrix} &\sim \mathcal{N}\left(
        \begin{bmatrix}
            0 \\ 0
        \end{bmatrix},
        \begin{bmatrix}
            \tau_{00} & \tau_{01} \\ \tau_{01} & \tau_{11}
        \end{bmatrix}
    \right).
\end{align}
The cross-level interaction coefficient $\gamma_{11}$ shows the level of interaction, or how much $Z$ moderates the slope $\beta_{1j}$.
We perform a Wald test~\citep{wald1943tests} on the null hypothesis $\gamma_{11} = 0$ and report the corresponding $p$-value. 
Also we summarize the magnitude of moderation by the proportion of variance $\tau_{11}$ explained by introducing $Z$~\citep{raudenbush2002hierarchical},
\begin{equation}
    R^2_\text{slope} = 1 - {\tau_{11} ^\text{(mod)}}/{\tau_{11}^\text{(omn)}}.
\end{equation}

\section{Theoretical Analysis on a Toy Model}

In this section, we provide a theoretical analysis on why the slope of quality to FID should depend on the geometric descriptors using a simple toy model.
FID is defined as the 2-Wasserstein distance between the Inception-v3 features of reference and generated image sets, where the distribution of features are assumed to be normal.
For two normal distributions $\mathcal{N}(\mu_r, \Sigma_r)$ and $\mathcal{N}(\mu_g, \Sigma_g)$, 2-Wasserstein distance has a closed form
\begin{equation}
    W_2^2 
    = \left\Vert \mu_r - \mu_g \right\Vert_2^2 + \mathrm{tr} \left(
        \Sigma_r + \Sigma_g - 2 \left(
            \Sigma_r^{1/2} \Sigma_g \Sigma_r^{1/2}
        \right)^{1/2}
    \right).
\end{equation}

\paragraph{Toy Model.}

We define reference and generated feature distributions for our toy model.
Let the reference features $\mathbf{x} \in \mathbb{R}^D$ follow a normal distribution
\begin{equation}
    \mathbf{x} \sim \mathcal{N}(0, P)
\end{equation} 
where $P = \mathrm{diag}(1, \dots, 1, 0, \dots, 0)$ is a projection matrix with $r$ ones and $D-r$ zeros. 
The features span $r$ linear dimensions on the feature space.
Then we obtain generated feature samples $\hat{\mathbf{x}} \in \mathbb{R}^D$ by adding noise to the reference features as
\begin{equation}
    \hat{\mathbf{x}} = \mathbf{x} + \epsilon, 
    \quad \epsilon \sim \mathcal{N}(0, \lambda ^2 I)
\end{equation}
where $\lambda > 0$ is the noise level and $I$ is the identity matrix.
Then the generated feature samples follow another normal distribution
\begin{equation}
    \hat{\mathbf{x}} \sim \mathcal{N} (0, P + \lambda^2 I).
\end{equation}
Using the definition of 2-Wasserstein distance
we have
\begin{align}
    W_2^2
    &= \mathrm{tr} \left(
        P + \left(P + \lambda^2 I \right) - 2\left( 
            P^{1/2} \left(P + \lambda^2 I \right) P^{1/2}
        \right)^{1/2}
    \right) \\
    &= \mathrm{tr} \left( 
        2P + \lambda^2 I - 2 \sqrt{1 + \lambda^2 } P
    \right) \\
    &= D\lambda^2 + 2r \left(
        1 - \sqrt{1 + \lambda ^2}
    \right) \\
    &= (D-r) \lambda^2 + \mathcal{O} \left( \lambda^4 \right).
\end{align}
Since $0 < r \leq D$, a large $r$ value reduces the magnitude of FID's response to a change in quality, and a small $r$ value amplifies it.

\paragraph{Geometric Descriptors. }

When the number of samples $n$ goes to infinity, we have
\begin{equation}
    \mathrm{erank}(A) = \mathrm{rank}(A) = r.
\end{equation}
Also, when $r$ is larger, the distance to the $k$-th nearest neighbor should increase since there are more linear dimensions in which the samples are spread, resulting in lower distributional density. 
Therefore the geometric descriptors govern the magnitude of FID's response to quality changes in this toy model.
Reproducing the empirical sign reversal on dispersed datasets, however, would require a richer noise model that captures anisotropic mismatch between the generated and reference covariances.
We leave this to future work.

\section{Future Work}

An important direction for future work is to test whether our findings generalize across different generative models.
In this paper, we use Stable Diffusion 1.5~\citep{rombach2022high} with the DDIM sampler~\citep{song2020denoising} to isolate the effect of the reference dataset.
However, the six reference datasets span markedly different domains, from single-domain face images to open-domain natural scenes.
As a result, the generator may fit these domains to different degrees, and such variation could in principle act as an unmeasured covariate alongside dataset geometry.
Because we are not aware of a single openly available generator that is competitive on both face synthesis and open-domain text-to-image generation, we adopt a widely benchmarked diffusion model as a representative testbed.
A natural extension is therefore to replicate our moderation analysis with other generator families and samplers, in order to test whether the dependence on reference geometry is specific to Stable Diffusion 1.5 with DDIM.

Another important direction for future work is to expand the set of reference datasets.
The literature on generative model evaluation relies on a relatively small pool of reference datasets, often inherited from prior benchmarks~\citep{dehghani2021benchmark, otani2023toward}.
We selected six datasets to span the concentrated-to-dispersed axis described in \cref{subsec:design}, which was sufficient to reject the omnibus null hypothesis and to explain a large fraction of the between-dataset slope variance using geometric descriptors.
With a larger and more diverse set of reference datasets, between-dataset variance components could be estimated more precisely, and additional candidate moderators could be tested with less risk of overfitting at the dataset level.

\section{Licenses}

\begin{itemize}
    \item \textbf{Stable Diffusion 1.5}
    - weights released under the CreativeML Open RAIL-M license (v1.0; \url{https://github.com/CompVis/stable-diffusion/blob/main/LICENSE})
    \item \textbf{FID} - clean-FID implementation by Parmar et al., released under the MIT License (v1.0; \url{https://github.com/GaParmar/clean-fid/blob/main/LICENSE})
    \item \textbf{FFHQ}:
    \begin{itemize}
      \item Dataset materials (metadata, scripts, documentation) released under the Creative Commons Attribution-NonCommercial-ShareAlike 4.0 International license (CC BY-NC-SA 4.0; \url{https://creativecommons.org/licenses/by-nc-sa/4.0/}) by NVIDIA Corporation.
      \item Individual images carry their original Flickr-author licenses (a mix of CC BY 2.0, CC BY-NC 2.0, Public Domain Mark 1.0, CC0 1.0, and U.S. Government Works); per-image license is recorded in the dataset metadata (\url{https://github.com/NVlabs/ffhq-dataset/blob/master/LICENSE.txt}).
    \end{itemize}
    \item \textbf{CelebA-HQ} - released for non-commercial research purposes only under the CelebA dataset terms by MMLab, CUHK (\url{https://mmlab.ie.cuhk.edu.hk/projects/CelebA.html}).
    \item \textbf{MJHQ-30K} - released by Playground AI on HuggingFace without an explicit dataset-level license (\url{https://huggingface.co/datasets/playgroundai/MJHQ-30K}); underlying images are Midjourney generations governed by the Midjourney Terms of Service (\url{https://docs.midjourney.com/hc/en-us/articles/32083055291277-Terms-of-Service}).
    \item \textbf{ImageNet} - released for non-commercial research and educational use under the ImageNet Terms of Access by Princeton University and Stanford University (\url{https://image-net.org/accessagreement}).
    \item \textbf{Flickr30K}:
    \begin{itemize}
      \item Annotations released for non-commercial research and educational use by Hockenmaier et al. (\url{https://shannon.cs.illinois.edu/DenotationGraph/}).
      \item Underlying images governed by Flickr Terms of Use; users must comply with Flickr's rules when reusing or redistributing any Flickr30K images.
    \end{itemize}
    \item \textbf{MS COCO 2014}:
    \begin{itemize}
      \item Annotations released under the Creative Commons Attribution 4.0 International license (CC BY 4.0; \url{https://creativecommons.org/licenses/by/4.0/})  
      \item Underlying images governed by Flickr Terms of Use; users must comply with Flickr's rules when reusing or redistributing any COCO images. 
    \end{itemize}
\end{itemize}


\end{document}